\documentclass[runningheads,a4paper]{llncs}

\usepackage{amssymb}
\usepackage{amsmath}
\usepackage{graphicx}
\usepackage{verbatim}
\usepackage{multirow}
\usepackage[table,xcdraw]{xcolor}
\usepackage{caption}
\usepackage[export]{adjustbox}

\usepackage{url}
\urldef{\mailsa}\path|{hekim, swkim, jhlee}@lunit.io|

\newcolumntype{Y}{>{\centering\arraybackslash}X}

\begin{document}

\mainmatter  

\title{Keep and Learn: Continual Learning by Constraining the Latent Space for Knowledge Preservation in Neural Networks}
\titlerunning{Keep and Learn: Knowledge Preservation in Neural Networks}

%
%
\author{Hyo-Eun Kim\thanks{Corresponding Author: hekim@lunit.io}\and Seungwook Kim\and Jaehwan Lee}
\authorrunning{H.-E. Kim, S. Kim, and J. Lee}
\institute{Lunit Inc., Seoul, South Korea}


%
%


\maketitle

%

\begin{abstract}
Data is one of the most important factors in machine learning. However, even if we have high-quality data, there is a situation in which access to the data is restricted. For example, access to the medical data from outside is strictly limited due to the privacy issues. In this case, we have to learn a model sequentially only with the data accessible in the corresponding stage. In this work, we propose a new method for preserving learned knowledge by modeling the high-level feature space and the output space to be mutually informative, and constraining feature vectors to lie in the modeled space during training. The proposed method is easy to implement as it can be applied by simply adding a reconstruction loss to an objective function. We evaluate the proposed method on CIFAR-10/100 and a chest X-ray dataset, and show benefits in terms of knowledge preservation compared to previous approaches.
\end{abstract}

\section{Introduction}

In a restricted multi-center learning environment where each chunk of data is only available at the corresponding center, we should learn a model incrementally without previous data chunks. Consider the scenario in which privacy-sensitive medical data are spread across multiple hospitals such that a machine learning model has to be learned sequentially.
If all data are available to be used concurrently, learning just with state-of-the-art deep learning models such as ResNet for image recognition~\cite{r07_resnet_cvpr2016} or GNMT for machine translation~\cite{r08_gnmt_arxiv2016} can be a good solution. However, if a data chunk from one stage is not available anymore in the following learning stages, it is hard to preserve the knowledge learned from the old data chunk because of the phenomenon known as catastrophic forgetting~\cite{r02_catastrophic_iclr2014}. This becomes more problematic especially in neural networks optimized with gradient descent~\cite{r01_catastrophic_psy1989}.

Overcoming catastrophic forgetting is one of the key research topics in deep learning. One naive approach is to fine-tune (FT) the model with the data accessible at each stage by learning from the up-to-date model parameters~\cite{r09_ft_cvpr2014}. Learning without Forgetting (LwF) is a representative method for overcoming catastrophic forgetting in neural networks~\cite{r10_lwf_eccv2016}. Before starting training in the current stage, output logits (LwF-logits) of the current training examples are calculated first, so that each example is paired with its true label and also the pre-calculated LwF-logit. The LwF-logits are used as pseudo labels for preserving old knowledge. Elastic Weight Consolidation (EWC) maintains old knowledge by constraining important weights (i.e. model parameters) not to vary too much~\cite{r11_ewc_nas2017}. The relative importance between weights is defined based on Fisher information matrix. Deep Generative Replay (GR)~\cite{r12_gr_arxiv2017} uses a generative adversarial network~\cite{r18_gan_nips2014}. GR learns a generative model and a task solving model at the same time, and the learned generator is used for sampling old data during current learning stage. The concept of GR is interesting, but samples from generative models are not suitable for use in certain applications such as medical imaging where pixel-level details include important radiographic features for diagnosis.

LwF and EWC are representative approaches for preventing catastrophic forgetting in neural networks based on two distinctive philosophies: controlling the output activation (LwF) or the model parameters (EWC). In this work, we preserve knowledge by modeling the feature space directly.\footnote{We denote \textit{feature space} to be the space of feature vectors, usually from the layer before the output layer. \cite{r10_lwf_eccv2016} showed that using the LwF-vectors of the second last hidden layer instead of the LwF-logits of the output layer had no benefit.} Based on the assumption that there exists better feature space for knowledge preservation, we model the high-level feature space and the output (logit) space to be mutually informative each other, and constrain the feature space to be in the modeled space during training. With experimental validation, we show that the proposed method preserves more knowledge than previous approaches.

\section{Baseline models}

\begin{figure*}[t]
\begin{center}
\includegraphics[width=1.0\textwidth]{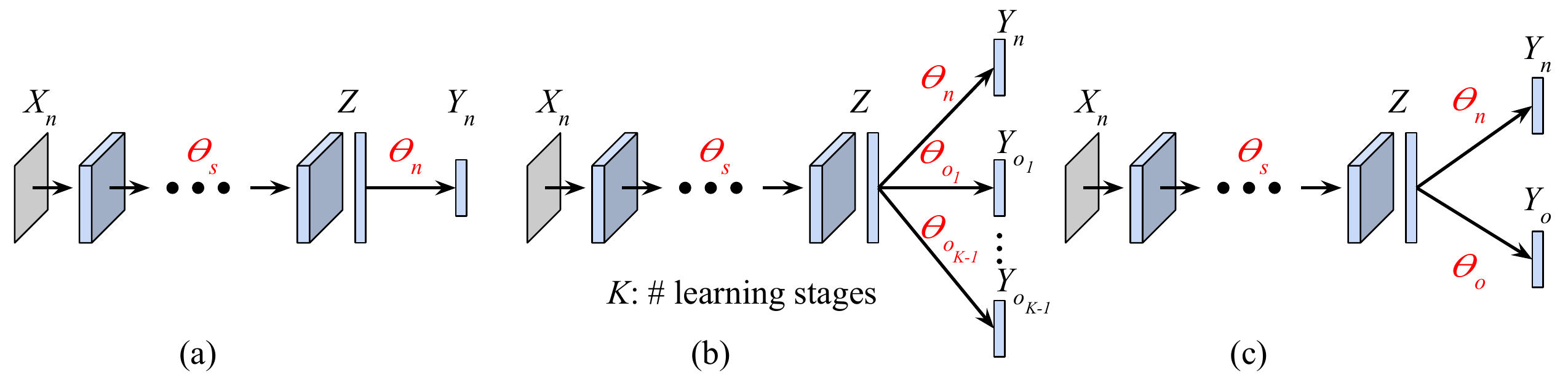}
\caption{Model architectures: (a) FT/EWC, (b) LwF, and (c) modified LwF (LwF+).}
\label{fig1:baseline}
\end{center}
\end{figure*}

LwF and EWC are originally proposed for preventing catastrophic forgetting in \textit{multi-task learning} where each task has its own data and the data used in previous tasks are not available when solving the current task. We call this as multi-center multi-task learning. We focus on \textit{multi-center single-task learning} where the model is learned with different data-chunk of the same task and access to each data-chunk is restricted. In this section, we define several baseline models for the multi-center single-task learning environment.
\\\\
\noindent\textbf{Fine-tuning (FT)} trains a model incrementally based on the model parameters learned in the previous stage. Figure~\ref{fig1:baseline}(a) shows the model architecture for FT. $X_n$, $Z$, and $Y_n$ are random variables for the input, latent, and output spaces, respectively. Target loss function $L_n(\theta)$ (e.g., negative-log-likelihood for classification) optimizes the model parameters $\theta$ which consist of $\theta_s$ (shared) and $\theta_n$ (new). In the first stage, $\theta$ is randomly initialized. In the following stages, $\theta$ is restored from the model learned in the previous stage.
\\\\
\noindent\textbf{Learning without Forgetting (LwF)} trains a model using both ground-truth labels and pseudo labels (pre-calculated LwF-logits). Figure~\ref{fig1:baseline}(b) demonstrates the $K$-th learning stage. $Y_n$ and $Y_{o_i}$ are the model's output for the current and the $i$-th stages for i in $\{1, ... , K-1\}$. The loss function is described as,
\begin{equation} \label{eq1_lwf}
	L(\theta) = L_n(\theta) + L_{LwF}(\theta), ~~~~ L_{LwF}(\theta) = \sum_i \lambda_{LwF} L_{o_i}(\theta),
\end{equation}
\noindent where $L_n(\theta)$ is the loss between the model output $y_n \in Y_n$ and its ground-truth label. $L_{o_i}(\theta)$ is the loss between the model output $y_{o_i} \in Y_{o_i}$ and its LwF-logit, and $\lambda_{LwF}$ is a weighting constant. $\theta_s$ and $\theta_n$ are initialized randomly in the first stage and restored from the previous stage in the following stages. In the $K$-th stage, $\theta_{o_{K-1}}$ is initialized with $\theta_n$ of the ($K-1$)-th stage and fine-tuned until the final stage. In the third stage, for example, $\theta_{o_1}$ and $\theta_{o_2}$ are restored from $\theta_{o_1}$ and $\theta_n$ of the second stage, respectively. For classification tasks, $L_n(\theta)$ and $L_{o_i}(\theta)$ are typically the cross-entropy loss.

In the multi-center multi-task learning environment, LwF preserves old knowledge by constraining the outputs of the old task-specific layers with corresponding pseudo labels. But, finding out the optimal feature space in terms of all the tasks becomes hard as the number of tasks (i.e. output branches) increases. 
\\\\
\noindent\textbf{Modified LwF (LwF+)}: LwF can be modified for the multi-center single-task learning. All the previous task-specific layers are merged into a single knowledge-preserving layer as shown in Figure~\ref{fig1:baseline}(c). So the loss function becomes,
\begin{equation} \label{eq2_lwf+}
	L(\theta) = L_n(\theta) + L_{LwF+}(\theta), ~~~~ L_{LwF+}(\theta) = \lambda_{LwF+} L_{o}(\theta),
\end{equation}
\noindent where $L_{o}(\theta)$ is the loss between $y_o \in Y_o$ and its pseudo label (LwF-logit). $\theta_s$ and $\theta_n$ are initialized randomly in the first stage and restored from the previous model in the following stages. $\theta_o$ is initialized with $\theta_n$ from the first stage and fine-tuned until the end of the learning stages.
\\\\
\noindent\textbf{Elastic Weight Consolidation (EWC)} constrains the model parameters by defining the importance of weights. Each parameter has its own weight-decay constant; the more important a parameter is, the larger the weight-decay constant. Based on the model in Figure~\ref{fig1:baseline}(a), the loss function is,
\begin{equation} \label{eq3_ewc}
	L(\theta) = L_n(\theta) + L_{EWC}(\theta), ~~~~ L_{EWC}(\theta) = \sum_j \frac{\lambda_{EWC}}{2} F_j(\theta_j - \theta_{p,j}^*)^2,
\end{equation}
\noindent where $\theta_{p,j}^*$ is the $j$-th model parameter learned in the previous stage and $F_j$ is the $j$-th element of the diagonal of the Fisher matrix $F$ for weighting the $j$-th model parameter $\theta_j$. $\lambda_{EWC}$ is a weighting constant. $\theta_s, \theta_n$ are randomly initialized in the first stage and restored from the previous model for the following stages.
\\\\
\noindent\textbf{EWCLwF (EWCLwF+)} is the combined model of EWC and LwF (LwF+). Since both methods keep old knowledge based on two distinctive approaches, they can be used complementarily. Based on the model architecture described in Figure~\ref{fig1:baseline}(b) with the loss function in Eq.~(\ref{eq1_lwf}), $L_{EWC}(\theta)$ in Eq.~(\ref{eq3_ewc}) is merged so the loss function becomes $L(\theta) = L_n(\theta) + L_{LwF}(\theta) + L_{EWC}(\theta)$. EWCLwF+ is similar to EWCLwF. Based on the model LwF+ in Figure~\ref{fig1:baseline}(c) with the loss in Eq.~(\ref{eq2_lwf+}), target loss becomes $L(\theta) = L_n(\theta) + L_{LwF+}(\theta) + L_{EWC}(\theta)$.

All the presented models are originated from the two representative methods for knowledge preservation in neural networks. Details of the experimental set-up for the baseline models will be explained in Section 4.

\section{Proposed Methodology}

In a general neural network model as in Figure~\ref{fig1:baseline}(a), the output $Y_n$ of the input data $X_n$ is compared with its true label, and the error is propagated backward from top to bottom, which encourages the latent variable $Z$ to be task-specific. To keep the previously learned knowledge, the latent space $Z$ should be informative enough to include the information of the input $X_n$.

\begin{figure}[t]
\centering
\begin{minipage}{.48\textwidth}
  \centering
  \includegraphics[width=1.0\linewidth, left]{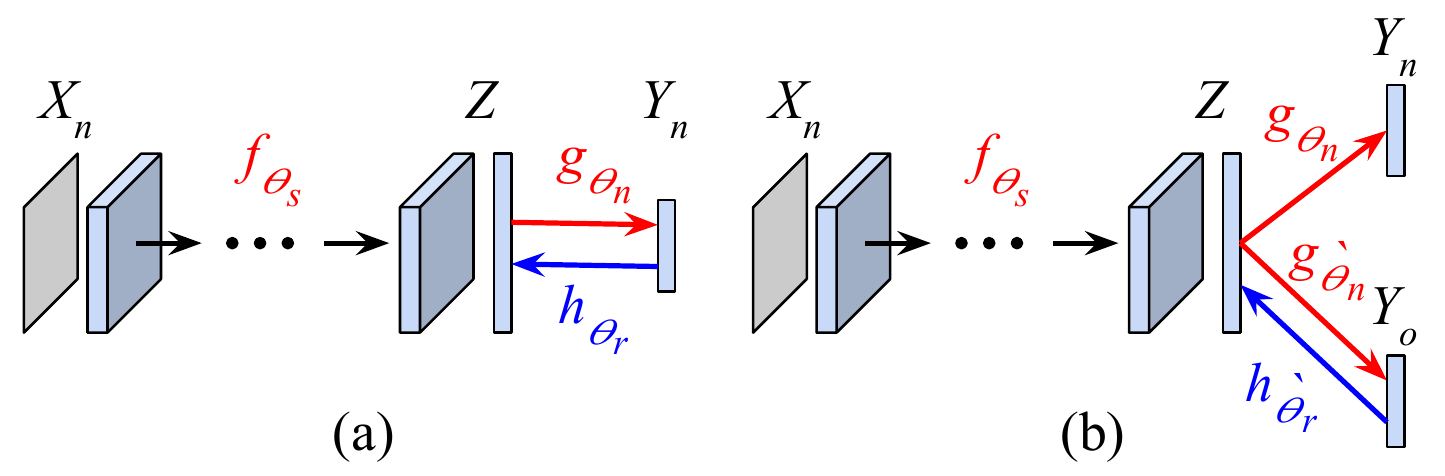}
  \caption{Proposed model architecture: (a) the first learning stage and (b) the following learning stages.}
  \label{fig2:proposed}
\end{minipage}%
\begin{minipage}{.02\textwidth}
\end{minipage}
\begin{minipage}{.02\textwidth}
\end{minipage}
\begin{minipage}{.48\textwidth}
  \centering
  \includegraphics[width=1.0\columnwidth, right]{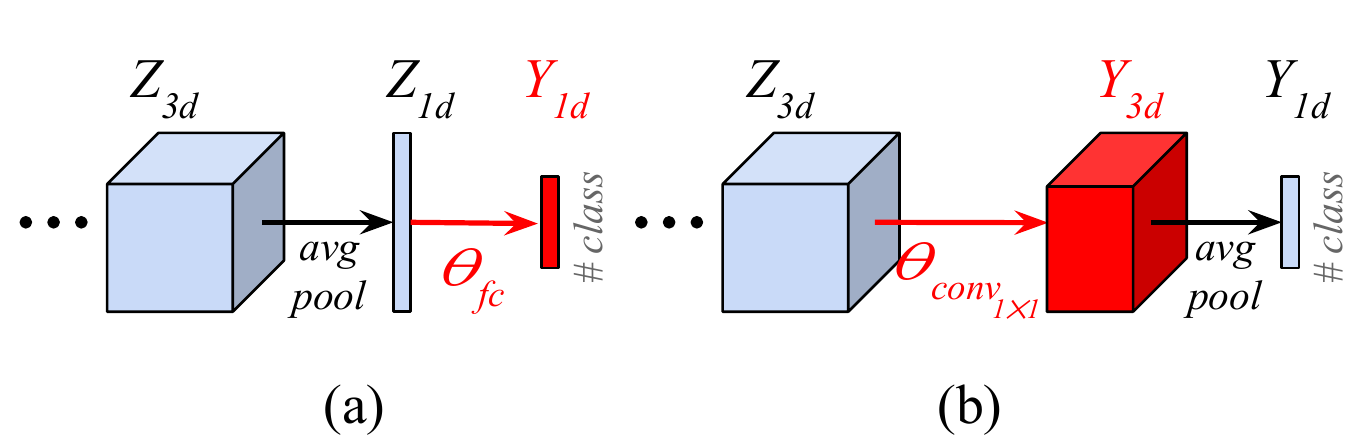}
  \caption{Top layers of ResNet: based on (a) $fc$ layer or (b) $conv_{1\times1}$ layer. Both are functionally equivalent.}
  \label{fig3:resnet_1x1}
\end{minipage}
\end{figure}

During learning the feature extractor $f$ of $\theta_s$ and the classifier $g$ of $\theta_n$, inverse function $h$ of $g$ ($h=g^{-1}$) can be approximately modeled by minimizing the $L2$ distance between the latent vector $z \in Z$ and its reconstruction $h(g(z))$ like Figure~\ref{fig2:proposed}(a). Without any constraints, minimizing the reconstruction loss easily makes the latent space $Z$ to be trivial in terms of the information that $Z$ can represent such that $\textbf{H}(Z)$ which is an entropy of $Z$ is low. Since $Z$ should be informative enough to minimize the task solving loss $L_n(\theta)$, joint learning with both the reconstruction and task solving losses prevents $Z$ from being trivial. It is known that minimizing the conditional entropy $\textbf{H}(Z \vert Y_n)$ can be done by minimizing the reconstruction error of $Z$ under the auto-encoder framework~\cite{r13_sdae_jmlr2010}. And minimizing the task solving loss $L_n(\theta)$ keeps $\textbf{H}(Z)$ not to reduce too much. As a result, $Z$ and $Y_n$ are being mutually informative from the joint learning with the two losses.\footnote{Note that the mutual information between $Z$ and $Y_n$ is $\textbf{I}(Z; Y_n) = \textbf{H}(Z) - \textbf{H}(Z \vert Y_n)$.}

Figure~\ref{fig2:proposed} shows the proposed model architecture. In the first stage, $f$, $g$, and $h$ (respectively parameterized by $\theta_s$, $\theta_n$, and $\theta_r$; initialized randomly in the first stage) are learned by minimizing the task solving and reconstruction losses concurrently. In the next stage, the parameters $\theta_o$ and $\theta_r$ of the functions $g'$ and $h'$ are restored from the $\theta_n$ and $\theta_r$ of the first stage and fixed during the rest of the learning stages.\footnote{$\theta_o, \theta_r$ are used to restore the modeled space, so they do not need to be fine-tuned.} $Y_n$ and $Y_o$ are the outputs for solving the task with current data and preserving previously-learned knowledge, respectively. Based on the loss function for LwF+ in Eq.(\ref{eq2_lwf+}), target $Z$ space modeled in the first stage can be kept in the following stages by fixing $\theta_o$ of $g'$ and $\theta_r$ of $h'$ and guiding the output $Y_o$ with LwF-logits. The loss function is shown below,
\begin{equation} \label{eq7_proposed}
	L(\theta) = L_n(\theta) + L_{LwF+}(\theta) + L_{rec}(\theta), ~~~~ L_{rec}(\theta) = \lambda_{rec} L2(\theta),
\end{equation}
\noindent where $\lambda_{rec}$ is a weighting constant for the reconstruction loss. LwF-logits for $Y_o$ are calculated in the same manner as in LwF+. $\theta_s$ and $\theta_n$ in the second stage are initialized with the parameters learned from the first stage and fine-tuned using the data in the corresponding stages until the end of the learning process.

Since we bound the $Z$ space with the space modeled in the first stage and fix the $\theta_o, \theta_r$ and $Y_o$ (with LwF-logits), $f$ tries to pull the new data examples into the modeled space which is remembering the previous data examples.

\section{Experiments}

\begin{figure}[t]
\begin{center}
\includegraphics[width=0.8\columnwidth]{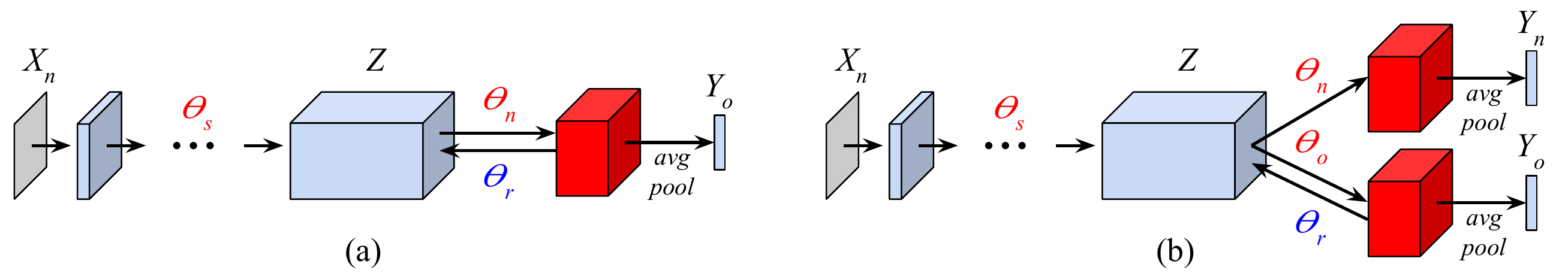}
\caption{Proposed model described in Figure~\ref{fig2:proposed} based on the modified ResNet in Figure~\ref{fig3:resnet_1x1}.}
\label{fig4:modified_resnet}
\end{center}
\end{figure}

We compare the proposed method with the baseline models in several image classification tasks. Base network is ResNet~\cite{r07_resnet_cvpr2016} which consists of multiple residual blocks and average-pooling (\textit{avgpool}) followed by a fully-connected (\textit{fc}) layer as shown in Figure~\ref{fig3:resnet_1x1}(a). The 3-D feature map $Z_{3d}$ extracted from the top-most residual block is pooled into a 1-D feature vector $Z_{1d}$ via \textit{avgpool}, and the output vector $Y_{1d}$ is obtained from $Z_{1d}$ through the final \textit{fc}. Given $z_{3d} \in Z_{3d}$ of an input example, $y_{1d} \in Y_{1d}$ is given by $g_{\theta_{fc}}(avgpool(z_{3d}))$, where $g$ is the \textit{fc} layer parameterized by $\theta_{fc}$. $g$ and \textit{avgpool} are commutative because \textit{avgpool} is a linear operation. Based on the modified model in Figure~\ref{fig3:resnet_1x1}(b), the output $y_{1d}$ can be described as $y_{1d} = avgpool(g_{\theta_{{conv}_{1\times1}}}(z_{3d}))$, where $g$ is now an 1$\times$1 convolution layer ($conv_{1\times1}$) parameterized by $\theta_{{conv}_{1\times1}}$. We used the modified ResNet in order to model the approximate inverse function $h$ accurately before \textit{avgpool}. Both are equivalent in terms of their function, but the modified model requires more computation than the original ResNet. The proposed network architecture is shown in Figure~\ref{fig4:modified_resnet}. $\theta_n$ and $\theta_o$ are the model parameters of $conv_{1\times1}$ layers which are the replacement of \textit{fc} layers in the original ResNet.

Three datasets are used for experimental validation; CIFAR-10/100~\cite{r14_cifar_UT2009} and chest X-rays (CXRs) for natural image and medical image classification. ResNet-56, 110, 21 are the base models for CIFAR-10, CIFAR-100, and CXRs, respectively. Each network consists of an initial convolution layer, three sets of $N$ consecutive residual blocks, and a final $conv_{1\times1}$ layer. In ResNet-21, an additional convolution layer (kernel 3$\times$3, filter width 32, stride 2) with maxpooling (kernel 2$\times$2, stride 2) is added as conv-bn-relu-maxpool (bn: batch normalization~\cite{r17_bn_icml2015}, relu: rectified linear unit~\cite{r05_alexnet_nips2012}) before the initial convolution to expand receptive field for large-size CXRs. Table~\ref{table1:resnet} summarizes the layer components. The top layer of ResNet-21 is modified from its original architecture and this will be explained in Section 4.2. Approximate inverse function $h$ (of $g$) parameterized by $\theta_{r}$ in Figure~\ref{fig4:modified_resnet} consists of multiple consecutive convolutions. $h$ in ResNet-56, 110, 21 for CIFAR-10, 100, CXRs includes four, three, three consecutive 3$\times$3 (stride 1) convolution layers with filter widths (64, 128, 128, 64), (256, 256, 256), (32, 64, 128) followed by a single bn-relu, respectively.

\setlength{\tabcolsep}{5pt}
\begin{table}[t]
\caption{Layer components. $N$, $C$, $R$ are \# of residual blocks, a conv layer, a residual block, respectively; e.g., $R_1$ of ResNet-110 has 18 \# of two consecutive 3$\times$3 conv layers with filter width 64. Downsampling with stride 2 is performed by $R_2$ and $R_3$.}
\label{table1:resnet}
\def\arraystretch{1.3}
\begin{center}\tiny
	\begin{tabular}{ l | c c c c c c}
	\hline
							& $N$ 	& $C_{init}$		& $R_1$						& $R_2$							& $R_3$							& $C_{1\times1}$ 	\\
	\hline
	ResNet-56		& 9		& $3 \times 3, 16$			& $\left[3 \times 3, 16\right] \times 2$ 	& $\left[3 \times 3, 32\right] \times 2$		& $\left[3 \times 3, 64\right] \times 2$		& $1 \times 1, 10$		\\
	ResNet-110		& 18		& $3 \times 3, 16$			& $\left[3 \times 3, 64\right] \times 2$ 	& $\left[3 \times 3, 128\right] \times 2$		& $\left[3 \times 3, 256\right] \times 2$		& $1 \times 1, 100$	\\
	ResNet-21		& 3		& $3 \times 3, 32$			& $\left[3 \times 3, 32\right] \times 2$ 	& $\left[3 \times 3, 64\right] \times 2$		& $\left[3 \times 3, 128\right] \times 2$		& $1 \times 1, 2$		\\
	\hline
	\end{tabular}
\end{center}
\end{table}

For CIFAR-10/100, the initial learning rate of 0.1 is decayed by $\frac{1}{10}$ every 40 epochs until the 120-th epoch. For CXRs, the initial learning rate of 0.01 is decayed by $\frac{1}{10}$ every 20 epochs until the 80-th epoch. Weight decay constant of 0.0001 and stochastic gradient descent with momentum 0.9 are used. For CIFAR-10/100, 32$\times$32 image is randomly cropped from 40$\times$40 zero-padded image (4 pixels on each side of the original 32$\times$32 image) during training~\cite{r07_resnet_cvpr2016}. Each CXR is resized to 500$\times$500 and randomly cropped 448$\times$448 image is used for training. $\lambda_{EWC}$ for CIFAR-10, CIFAR-100, and CXRs are 0.1, 10.0, and 1.0, respectively. They are selected from the set $\{$0.1, 1.0, 10.0$\}$ by cross validation. $\lambda_{LwF}$ in Eq.~(\ref{eq1_lwf}) is $\frac{0.1}{K-1}$, where $K$ is the number of learning stages including the current one. $\lambda_{LwF+}$ and $\lambda_{rec}$ are 0.1 and 1.0. All experiments are done with tensorflow~\cite{r16_tf}.

\subsection{CIFAR-10/100}

CIFAR-10/100 have 10/100 classes with 32$\times$32 50k/10k training/test images, respectively. In our experiment, 10k training images are used for validation and the model which performs the best on the validation set is selected for evaluation on the test set. The remaining 40k training images are splitted into four sets (10k/set). Each model is trained continually in the multi-center single-task learning set-up, where each center has 10k training images and the task is 10/100-class classification. Table~\ref{table2:result_c10} shows the error rates on the test set with mean (std) of five trials. LwF+, EWCLwF+ mostly perform better than LwF, EWCLwF; i.e. LwF+, EWCLwF+ are more appropriate for the multi-center single-task learning. The proposed method performs the best as shown in this table. 

\setlength{\tabcolsep}{1.0pt}
\begin{table}[t]
\caption{CIFAR-10/100: test set (10k images) error rates - mean (std) of five trials.}
\label{table2:result_c10}
\def\arraystretch{1.3}
\begin{center}\tiny
	\begin{tabular}{ l | c c c c | c c c c}
	\hline
	\multirow{2}{*}{  }	& \multicolumn{4}{c |}{CIFAR-10}												& \multicolumn{4}{c}{CIFAR-100}					\\
						& stage-1				& stage-2				& stage-3				& stage-4				& stage-1				& stage-2				& stage-3				& stage-4			\\
	\hline
	FT				& 20.21(.151)	& 16.76(.419)	& 15.40(.174)	& 15.02(.174)	& 50.13(1.25)	& 42.79(.692)	& 40.53(.467)	& 38.96(.354)	\\
	EWC				& 19.87(.421)	& 16.70(.178)	& 15.42(.258)	& 14.77(.331)	& 49.93(.937)	& 42.52(.299)	& 40.72(.231)	& 38.94(.504)	\\
	LwF				& 20.28(.532)	& 16.62(.453)	& 15.46(.220)	& 14.68(.304)	& 50.41(.422)	& 42.70(.334)	& 39.50(.417)	& 37.51(.319)	\\
	LwF+			& 19.88(.574)	& 16.57(.194)	& 15.02(.238)	& 14.05(.115)	& 50.69(.760)	& 42.64(.887)	& 39.31(.490)	& 37.30(.558)	\\
	EWCLwF		& \textbf{19.79(.122)}	& 16.62(.041)	& 15.45(.413)	& 14.49(.183)	& 50.15(.552)	& 42.22(.481)	& 39.62(.338)	& 37.44(.526)	\\
	EWCLwF+		& 20.26(.474)	& 16.99(.410)	& 15.34(.440)	& 14.25(.239)	& 50.10(.439)	& 42.49(.335)	& 39.32(.288)	& 37.21(.377)	\\
	\hline
	Proposed		& 20.11(.431)	& \textbf{16.12(.253)}	& \textbf{14.54(.175)}	& \textbf{13.74(.195)}	& \textbf{49.87(.461)}	& \textbf{42.00(.479)	}& \textbf{38.81(.438)	}& \textbf{36.42(.373)}	\\
	\hline
	\end{tabular}
\end{center}
\end{table}

After stage-1, training data of the stage-1 (st-1-trn) is not used in the following stages anymore. So, we evaluate the final model with st-1-trn to see how much of st-1-trn has been forgotten after the final stage. For CIFAR-10, 85.75\%, 85.97\%, 88.64\%, 88.22\%, 89.40\% of st-1-trn are still preserved as correct at stage-4 for FT, EWC, LwF+, EWCLwF+, Proposed, respectively. For CIFAR-100, 58.67\%, 58.85\%, 65.57\%, 66.91\%, 69.34\% of st-1-trn are preserved correctly at the final stage (with the same ordering).

\subsection{Chest X-rays for Tuberculosis}

We experiment with a real-field medical dataset in order to verify the proposed method is also valid in a practical set-up. A total of 10,508 de-identified CXRs (from the Korean Institute of Tuberculosis~\cite{r15_tb_spie2016}) are used. It consists of 3,556 abnormal (tuberculosis; TB) and 6,952 normal cases. CXRs are commonly used for screening TB. The cases which require a follow-up test are recalled by radiologists. Among the 3,556 abnormal cases, 1,438 cases were diagnosed as active TB (TB-A) at the screening stage. The status of the remaining 2,118 cases which needed a follow-up sputum test could not be specified radiologically at the screening stage (TB-U). 80\% of the data are randomly selected for training and divided into four sets; 288(TB-A), 424(TB-U), 1390(Normal) per each set. The remaining 20\% are splitted evenly for validation and test; 143(TB-A), 211(TB-U), 696(Normal) for each set.

We modified the output layer of the model in order to exploit the status information of abnormality. Two output $conv_{1\times1}$ layers are used for 2-class (TB vs normal) and 3-class (TB-A, TB-U, and normal) classification, respectively. The 3-class $conv_{1\times1}$ is used for knowledge preservation. The 2-class $conv_{1\times1}$ is just for the performance measurement (AUC; area under ROC curve).

Table~\ref{table5:result_tb} summarizes AUC of each model with mean (std) of five trials. Except for the first stage, the proposed method  is always better than the others. The proposed method also performs the best in terms of the ensemble performance of the five trials; 0.9257, 0.9205, 0.9217, 0.9271, 0.9228, 0.9172, 0.9363 for FT, EWC, LwF, LwF+, EWCLwF, EWCLwF+, Proposed, respectively.  Figure~\ref{fig7:result_tb} is the ROC curves of the st-1-trn at stage-4 (similar to CIFAR-10/100), which implicitly shows that the proposed method is helpful to preserve old knowledge.

\setlength{\tabcolsep}{1.0pt}
\begin{table}[t]
\begin{minipage}[b]{0.56\linewidth}
	\def\arraystretch{1.8}
	\begin{center}\tiny
	\begin{tabular}{ l | c c c c}
	\hline
						& stage-1					& stage-2					& stage-3					& stage-4				\\
	\hline
	FT				& 0.811(.025)		& 0.842(.019)		& 0.882(.011)		& 0.892(.015)	\\	
	EWC				& 0.812(.016)		& 0.832(.025)		& 0.865(.012)		& 0.887(.008)	\\
	LwF				& 0.814(.020)		& 0.853(.026)		& 0.882(.020)		& 0.891(.019)	\\
	LwF+			& 0.806(.010)		& 0.844(.022)		& 0.881(.018)		& 0.898(.014)	\\
	EWCLwF		& \textbf{0.821(.019)}		& 0.841(.021)		& 0.869(.018)		& 0.890(.023)	\\
	EWCLwF+		& 0.817(.012)		& 0.852(.018)		& 0.871(.019)		& 0.884(.017)	\\
	\hline
	Proposed		& 0.813(.035)		& \textbf{0.869(.021)}		& \textbf{0.896(.017)}	& \textbf{0.909(.013)}	\\
	\hline
	\end{tabular}
	\end{center}
	\caption{CXRs for TB: test set AUC - mean (std) of five trials.}
	\label{table5:result_tb}
\end{minipage}\hfill
\begin{minipage}[b]{0.42\linewidth}
	\begin{center}
	\includegraphics[trim=10 0 30 0,clip,width=0.93\columnwidth]{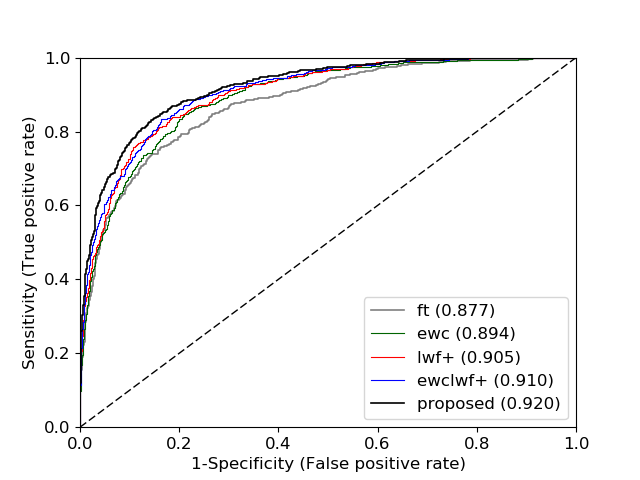}
	\captionof{figure}{ROC curves at stage-4 with stage-1 training data.}
    \label{fig7:result_tb}
	\end{center}
\end{minipage}
\end{table}

\section{Conclusion}

In this work, we raise the problem of catastrophic forgetting in multi-center single-task learning environment and propose a new way to preserve old knowledge in neural networks. By modeling the high-level feature space to be appropriate for knowledge preservation in the first stage and constraining the feature space to be in the modeled space during training in the following stages, we can preserve the knowledge learned in preceding stages. The proposed method is shown to be beneficial in terms of keeping the old knowledge in classification tasks. We need more experimental analysis beyond the classification such as lesion detection or segmentation, and we leave this for future work.

%

\bibliographystyle{splncs03}
\bibliography{CL}

\end{document}